\documentclass{article} 
\usepackage{nips13submit_e,times}
\usepackage{hyperref}
\usepackage{url}
\usepackage{graphicx}
\usepackage{amsfonts}
\usepackage{tabulary}
\DeclareGraphicsExtensions{.png,.pdf}

\title{CAESAR: Context Awareness Enabled \\ Summary-Attentive Reader}

\author{
Kshitiz Tripathi \thanks{The authors contributed equally.}\\
Yahoo Inc. \\
\texttt{kshitiz@stanford.edu} \\
\And
Long-Huei Chen \footnotemark[1] \thanks{Work done while author was a student at Computer Science Department, Stanford University, CA, USA.}\\
Graduate School of Interdisciplinary Information Studies  \\
The University of Tokyo\\
\texttt{longhuei@g.ecc.u-tokyo.ac.jp} \\
}

\nipsfinalcopy 

\begin{document}
\maketitle

\begin{abstract}
Comprehending meaning from natural language is a primary objective of Natural Language Processing (NLP), and text comprehension is the cornerstone for achieving this objective upon which all other problems like chat bots, language translation and others can be achieved. We report a Summary-Attentive Reader we designed to better emulate the human reading process, along with a dictiontary-based solution regarding out-of-vocabulary (OOV) words in the data, to generate answer based on machine comprehension of reading passages and question from the SQuAD benchmark. Our implementation of these features with two popular models (Match LSTM and Dynamic Coattention) was able to reach close to matching the results obtained from humans.
\end{abstract}

\section{Introduction}

Endowing machines with the ability to understand and comprehend meaning is one of the ultimate goals of language processing, one that holds promise to revolutionize the way people interact with and retrieve information from machines \cite{yin2016attention}. The goal was outlined by Richardson et al. with the MCTest dataset they proposed \cite{richardson2013mctest}, in which questions are provided for which the answer can only be found in the associated passage text. To perform well on such task, machine comprehension models are expected to possess some sort of semantic interpretation of text along with probabilistic inference to determine the most probable answer. 

Here we propose a novel approach to better question-answering by preprocessing document text before answer selection from the passage. This is achieved by a summarization engine that is applied to truncate document length to enable narrower focus on the correct answer span. The summarization takes into account the nature of the question, and therefore can help eliminate unrelated answer candidates in the passage that are likely to be erroneously selected. The summarization engine also assists in machine efficiency, since the encoder is more likely to focus directly around the answer span. The summarized document is then passed through a coattention-based encoder, and the encoded message is passed along to an answer pointer decoder to predict the answer.

\section{Related Work}

\subsection{Machine Comprehension and Question-Answering}

We begin with an overview of the features adopted by currently high-performing models evaluated on the SQuAD question-answering dataset. Instead of summarizing and repeating similar functionality shared among the implementations, we list characteristics that are unique to the models, from which we drew inspiration to develop our CAESAR question-answering engine.
\begin{itemize}

\item
\textbf{Multi-Perspective Context Matching} :
During preprocessing, character-level embedding are created from a LSTM RNN to assist with words with which pre-existing word embeddings are not available \cite{wang2016multi}. Then the document is filtered based on word-level cosine similarity between question and document. The key to the model is the MPCM layer, which compares the document with question in multiple perspectives that are applicable to different kinds of answer spans as seen in passage. 

\item
\textbf{Match-LSTM}:
In this model, bi-directional match-LSTM RNN created from concatenation of document and attention-weighted question is applied to predict the answer span \cite{wang2016machine}. The answer-pointer layer predict answer from the 2 directions of the LSTM bi-directional RNN hidden state. Sentinel vector is appended at the end of document to facilitate determination of answer span ends. 

\item
\textbf{Dyanmic Coattention Network} : In the dynamic coattention network, the coattention mechanism uniquely attends to both the question and the document simultaneously \cite{xiong2016dynamic}. In addition to the conventional attention of question in light of document words and the attention of the document in light of the question, the authors also calculated the summaries of attention in light of each word of the document. They also applied an iterative dynamic pointer decoder to predict on the answer span. Because the model alternates between predicting the start and the end point, recovery from local maxima is more likely.

\item
\textbf{Bi-directional Attention Flow} : 
The model also employed character-level embeddings with Convolutional Neural Networks \cite{seo2016bidirectional}, which are then combined with word embeddings via a Highway Network. Interestingly, in this model the attention flow is not summarized into single feature vectors, but is allowed to flow along with contextual embeddings through to subsequent modeling.

\item
\textbf{RaSoR} :
To address lexical overlap between question and document, for each word in document they created both passage-aligned and passage-independent question representation \cite{lee2016learning}. For passage-aligned representation, fixed-length representation of the question is created based on soft-alignments with the single passage word computed via neural attention. For passage-independent representation, the question word is compared to the universally learned embeddings instead of any particular word.

\item
\textbf{ReasoNet} :
The ReasoNet specifically mimics the inference process of a human reader by reading a document repeatedly, with attention on different parts each time until a satisfied answer is found \cite{shen2016reasonet}. The authors present an elegant approach for ReasoNet  to dynamically determine
whether to continue the comprehension process after digesting intermediate results, or to terminate reading when it concludes that existing information is adequate
to produce an answer. Since termination state is a discrete variable and non-trainable by canonical back-propogation, they use Reinforcement Learning to solve that.

\item
\textbf{Dynamic Chunk Reader} :
After obtaining attention vectors between each word in the question and the document, the dynamic chuck reader creates chunk representation that encodes the contextual information of candidate chunks selected from the document  \cite{yu2016end}. Then cosine similarities is evaluated between the chuck repreeation and the question, and the highest-scoring chuck is taken as the answer.

\item
\textbf{Fine-grained Gating}:
While other methods typically use word-level representation, or character-level representation, or both \cite{yang2016words}, in fine-grained gating a gate is applied to dynamically choose between the word or character-level representations for each document word. Word features such as named entity tags, part-of-speech tags, binned document frequency vectors, and the word-level representations all form the feature vector of the gate.

\end{itemize}

\section{Data Source}

Building on the basis of similar evaluation metrics of machine comprehension, Rajpurkar et al. released the Stanford Question Answering dataset (SQuAD) \cite{rajpurkar2016squad}, which has quickly become the de facto standard of comparison among machine comprehension models \cite{neelakantan2016learning}. The reason of the dataset's wide applicability is several-fold: first, the dataset is at least 2-orders of magnitude larger from other similar releases. Secondly, the questions are human-curated and are realistic approximation of real-life needs. Thirdly, the corresponding answers to each question are of a variety of nature and can test the generalizability of machine comprehension. Finally, the answer can be an arbitrary span instead of one of the pre-determined choices. SQuAD is comprised of around 100K question-answer pairs, along with a context paragraph. The context paragraphs were extracted from a set of articles from Wikipedia. Humans generated questions using that paragraph as a context, and selected a span from the same paragraph as the target answer.

\begin{table*}[!t]
    \centering
    \begin{tabulary}{\linewidth}{|C|L|}
    \hline 
     Document & Imperialism is a type of advocacy of empire. Its name originated from the Latin word "imperium", which means to rule over large territories. Imperialism is "a policy of extending a country's power and influence through colonization, use of military force, or other means". Imperialism has greatly shaped the contemporary world. It has also allowed for the rapid spread of technologies and ideas. \underline{The term imperialism has been} \underline{applied to Western (and Japanese) political and dominance especially in Asia and Africa} \underline{in the 19th and 20th centuries.} Its precise meaning continues to be debated by scholars. Some writers, such as Edward Said, use the term more broadly to describe any system of domination and subordination organised with an imperial center and a periphery. \\ \hline 
     Question & \textit{The term imperialism has been applied to western countries, and which eastern county?} \\ \hline 
     Answer & \textbf{Japan} \\ \hline 
    \end{tabulary}
    \caption{Example from the SQuAD dataset. Provided with a paragraph of text, we built an automated system to generate response based on a query by taking a sentence/phrase from the paragraph.}
    \label{tab:example}
\end{table*}

\section{Match-LSTM with Answer Pointer Decoder Baseline}

In Match LSTM model, the authors present a layered network architecture consisting of three layers.

We begin with an LSTM layer that preprocesses the passage and the question using LSTMs. A preprocessing layer uses standard one dimension LSTM to process the embedded passage and the question and collect the entire sequence of hidden states to generate Document and Question representations, $H^p$ and $H^q$, respectively.
$$H^p = \mathrm{\vec{LSTM}}(P), H^q = \mathrm{\vec{LSTM}}(Q)$$

The LSTM preprocessing helps incorporate context from the respective passages, and enable the encoded hidden state to include information contained words that are adjacent to the word in question. Also note that we shared a single LSTM unit for both the question and the document passage, a decision made for the purpose of shared parameters training, and is unlike the original Match-LSTM paper.

Afterwhich, we add a match-LSTM layer that tries to match the passage against the question. This is the core of the network where attention vector $\alpha_i$ is computed by globally attending to the hidden representation of each word of the passage for the entire question.
$$\vec{G}_i = \mathrm{tanh}(W^qH^q + (W^ph_{i}^p + b^p) \otimes e^Q)$$
$$\vec{\alpha}_{i} = \mathrm{softmax}(w^T\vec{G}_i + b \otimes e^Q)$$
where $W^p$, $W^q, W^r\in R^{lXl}$ $b^p, w \in R^l$ and $b \in R$. $h_i$ is the hidden state for the one-directional match-LSTM which receives the the concatenation of passage representation and attention weighted question representation. A similar LSTM is used for representation in reverse direction. Omitting the detail as they exactly match the Match-LSTM paper. Finally, the hidden states from both the LSTM are concatenated which is passed to the next layer for inference.

An Answer Pointer (Ans-Ptr) is located at the top layer is motivated by the Pointer Net introduced by Vinyals et al. \cite{NIPS2015_5866}. The authors mentions two variants for this layer: predicting the entire answer Sequence vs predicting only the answer span. The initial testing showed us that boundary model was better than the sequence model, which was also inline with the results mentioned in the paper, so we continued with the boundary model which was also simpler to implement.
 
 
\begin{figure}[t!]
\begin{center}
  \includegraphics[width=\textwidth]{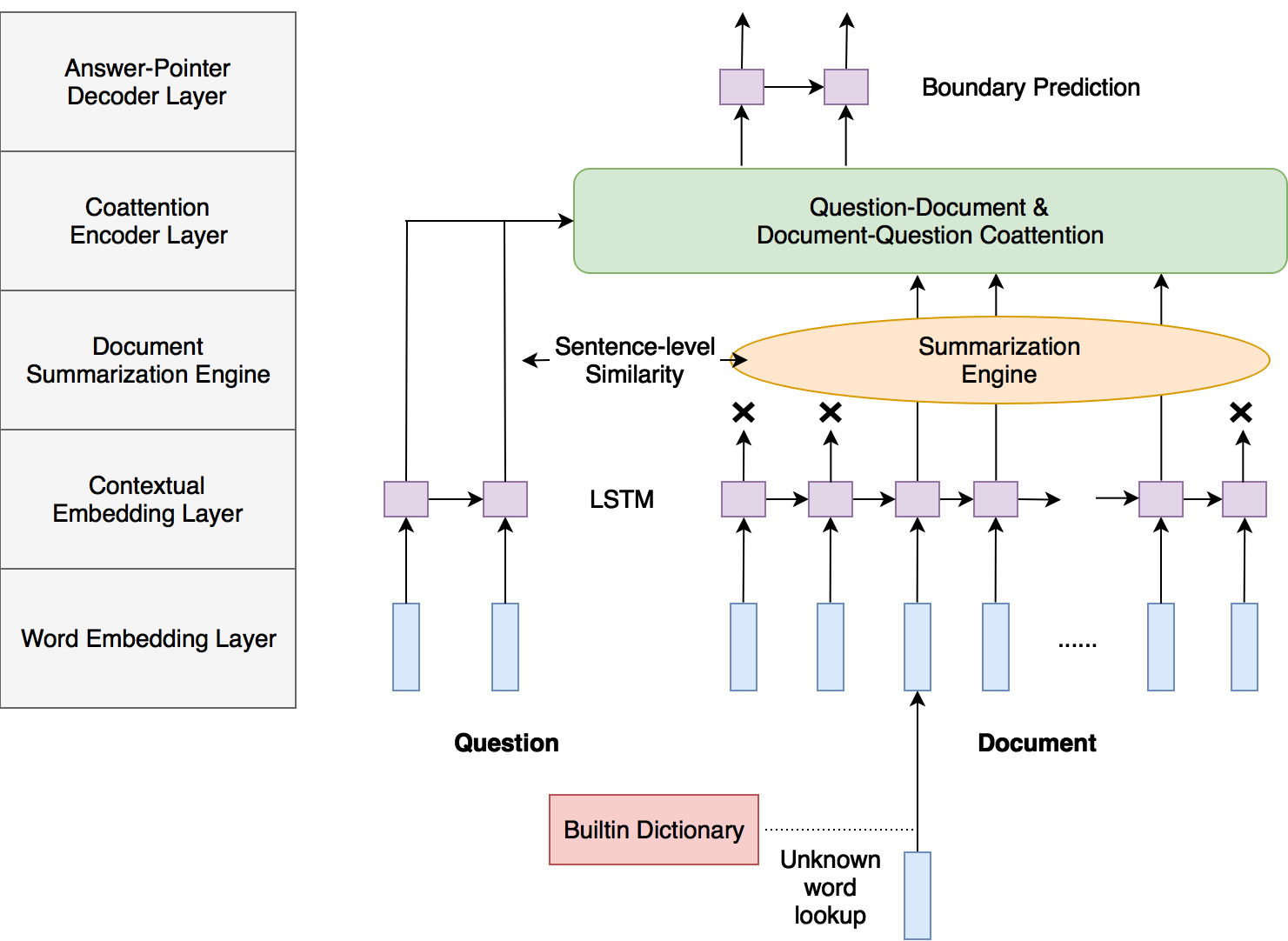}
\caption{Overall Network Architecture}
  \label{fig:architecture}
  \end{center}
\end{figure}

\section{Dynamic Coattention}
After setting up a baseline with the Match LSTM model, we started to implement Dynamic Coattention Model \cite{xiong2016dynamic} which had an advanced attention mechanism and a novel iterative approach for finding the answer span. Similar to Match-LSTM, this model also consisted of layers describe below.

\subsection{Document and Question Encode Layer} \label{coattention_preprocessing}
This layer is similar to the preprocessing layer of Match-LSTM model but with two modifications: the question and context tokens are fed to the same LSTM unit to obtain their encoding. And, there is an additional linear layer on top op the question encoding which results in general scoring function \cite{DBLP:journals/corr/LuongPM15} instead of simple dot product between question and document encoding for attention weights calculation. After this is completed, we obtain two matrices $D \in \mathbb{R}^{m \times l}$ and $Q \in \mathbb{R}^{n \times l}$. Each of the same dimension as the previous layer but now with contextual information. Also a nonlinear layer is placed upon the question encoding to produce $Q'$ from $Q$, to introduce variation between the question and document encoding space \unskip~\cite{wang2016machine}.

\subsubsection{Unknown Word Lookup}
A major bottleneck of the model in case of test/dev prediction was the presence of many new words in the test/dev dataset which were missing in the vocabulary constituted purely of training dataset. Since our network performed better with constant embeddings, we had the flexibility of enhancing the vocabulary before test/dev prediction. And upon including the Unknown Word Look-up module, we got a lift of 6-9\% in each F1 and EM score which was very crucial for the success of our model. We used an efficient hash join technique for the look-up and on test servers, it took just 60-100sec for looking up the missing word in the 2.19M Glove 840B 100D dataset which gave us more iterations to test.

\subsection{Coattention Encoder Layer}

The coattention mechanism attends to the question and document simultaneously, and finally fuses both attention contexts \unskip~\cite{xiong2016dynamic}, which proceeds as follows:

First, the affinity matrix calculates scores corresponding to all pairs of document words and question words
$$L = Q'D'^T \in \mathbb{R}^{n \times o}$$
then attention weights are produced by normalizing the affinity matrix. Then the affinity matrix is normalized column-wise to produce the attention weights across the document for each word in the question
$$A^Q = \mathrm{softmax} \left( L \right) \in \mathbb{R}^{n \times o}$$
and normalized row-wise to produce the attention weights across the question for each word in the document, $$A^D=\mathrm{softmax}\left(L^T \right)\in\mathbb{R}^{o\times n}$$

We then obtain summaries, or attention contexts, of the document in light of each word of the question \cite{seo2016bidirectional}.
$$C^Q = A^QD' \in \mathbb{R}^{ n \times l }$$

$A^DQ$ is summaries of the question in light of each word of the documents, and $A^DC^Q$ is the summaries of the previous attention contexts in light of each word of the document. These two operations are done in parallel, the latter can be interpreted as the mapping of question encoding into space of document encoding.
$$C^D=A^D\lbrack Q',C^Q\rbrack\in\mathbb{R}^{o\times2l}$$

We define $C^D$as a co-dependent representation of the question and document, as the coattention context. The last step is the fusion of temporal information to the coattention context via a bidirectional-LSTM.
$$U = \mathrm{Bi\mathrm{-}LSTM} \left( [D', C^{D}]  \right) \in \mathbb{R}^{o \times 2l}$$

\subsection{Summary-Attentive Reader}

\begin{figure}[!htb]
\centering
\includegraphics[width=\textwidth]{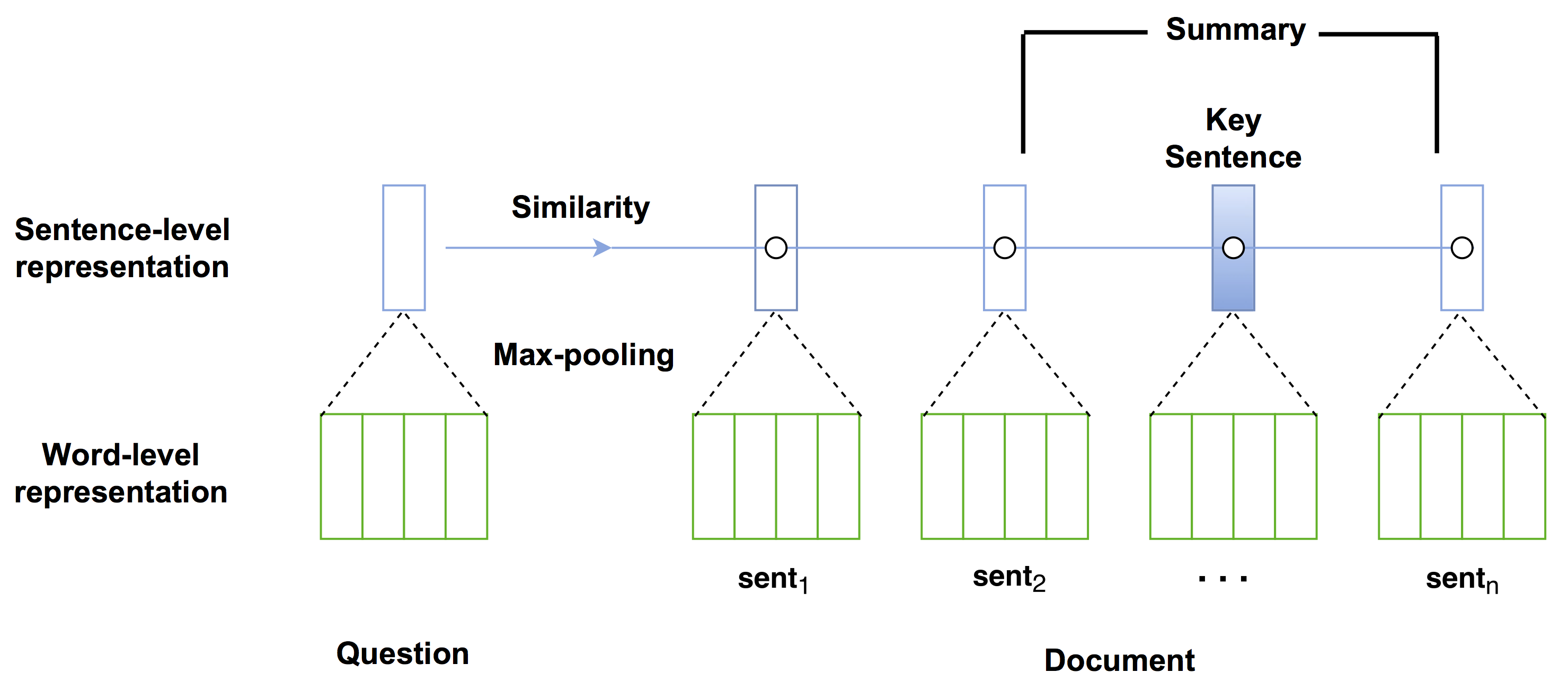}
\caption{The summarization engine first selects the key sentence in the document, which has the highest similarity to the question. Then the document is truncated around the key sentence to obtain the summary.}
\label{fig:sentencerep}
\hfill
\end{figure}

To facilitate better answer prediction, we created a summarization engine that can preprocess the document to generate a shortened summary that is more likely to contain the answer. This is motivated by the human readers' approach to question answering: when we are looking for an answer to some question from a paragraph, we would rarely read the article word-by-word in its entirety before selecting an answer. Instead we would often rapidly scan the document passage, identify sentences containing similar wording to the question, then select the answer based on our understanding of the text.

In an effort to emulate the aforementioned human question-answering process, we designed the summarization engine to truncate the document into a shorter paragraph before further processing. This is carried out through the following steps (Fig. \ref{fig:sentencerep}):

\begin{enumerate}
    \item A sentence-tokenizer is applied to the document to identify span indices of the sentences.
    \item Max-pooling or mean-pooling is applied across all word embeddings of each sentence in the document, in order to obtain sentence-level representations of all sentences.
    \item Similar pooling techniques is applied to the question.
    \item The sentence-level representation of each sentence in document is compared with the question, in order to identify the sentence that has the highest cosine similarity as the key sentence.
    \item The document is truncated around the key sentence to the pre-specified summary length.
\end{enumerate}

After being shortened, the summarized document is allowed to flow through to the subsequent layers. Our summarized method ensures that similar attention mechanisms and prediction methods can be applied regardless of whether the document is full-length or has being shortened, since the word ordering and sentence meaning are preserved in this summarization engine.

\subsection{Decoder Layer}

The \textbf{Pointer Network} represents an effort in selecting specific locations within the source text as an answer produced in the output. The idea is actually quite elegant: rather than producing an attention vector based on weights calculated from a nonlinear transformation of the decoder hidden states along with all the encoder hidden states, we simply use the weight as a probability that the particular location in the encoder would be chosen. Even though the idea is, at first glance, simple, it has been successfully applied to a variety of tasks such as finding convex hulls, Delaunay triangulations, and the Travelling Salesman Problem \cite{NIPS2015_5866}.

In our specific use case, the attention mechanism is used again to obtain an attention weight vector $B \in \mathbb{R}^{2 \times o}$, where $b_{1, p}$ is the probability that the $p$-th word from the document is the start word of the answer, and $b_{2,q}$ is the probability that the $q$-th word is the end word.

\section{Experiment}

\subsection{Match-LSTM Baseline}

After many iterations, we achieved a significant score of ( F1=53\%, EM=39\%) (complete score in \ref{result_table} ) and settled for that as a baseline score. Although there was huge scope for improvement, but we wanted to explore other advanced models.

\subsection{Summary-attentive reader}

To evaluate the summarization engine, we truncate each document in the train set to different word length. The success of the engine is measured by the ground truth answer retain rate, which is the percentage of ground truth that are wholly contained in the summary we selected. This can serve as a metric of how well the engine is doing as we can expect the same summarization, when applied to document with unknown answer from the test set, will also retain the actual answer for subsequent layer prediction. 

\begin{figure*}[h!]
\begin{center}
\includegraphics[width=0.75\textwidth]{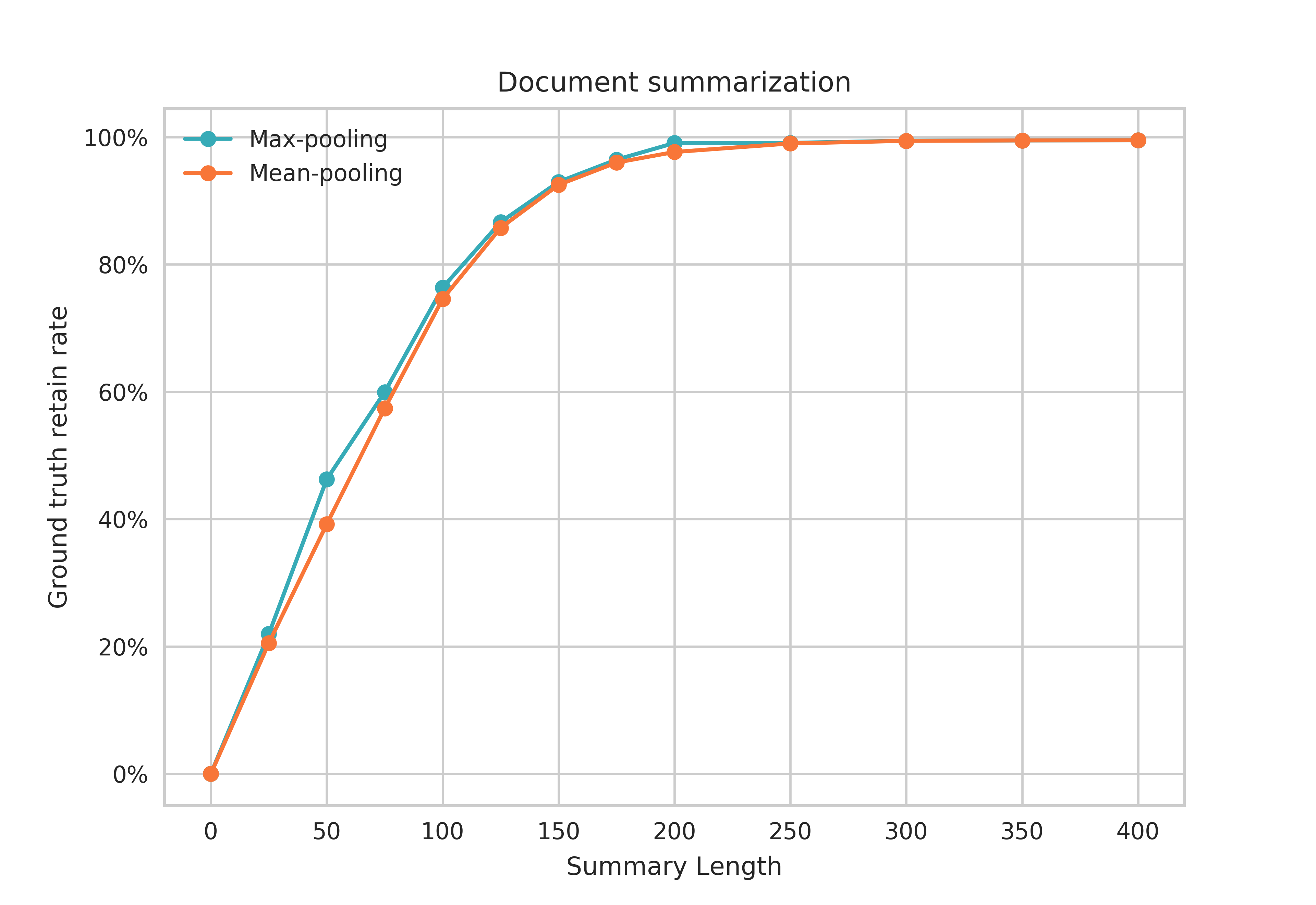}
\caption{Ground truth answer retain rate with respect to the truncated document summary length. The original document has a maximum length of 600, and the summarization engine is able to retain more than 92\% of ground truth answers in the summary when we truncate the document to length 150.}
\label{fig:summary}
\end{center}
\end{figure*}

Even though the maximum document size of our train set is 600 words, the summarization engine is able to retain more than 98.31\% of the ground truth answer when we truncate the document to 200 words or more (Fig. \ref{fig:summary}). Even when we truncate the document down to 150 words, 92.17\% of answers are contained in the summary. We also see that max-pooling performs marginally well than mean-pooling in the averaging across word embedding when obtaining sentence-level representation.

\subsection{Experiment Details}

Aftering experimenting with hyperparameters and various preprocessing settings, we settle on the following experiment details which gave the optimal result.

We apply pretrained GloVe word embeddings trained on common crawl with 840B tokens and 300 dimensions \cite{pennington2014glove}. The Stanford CoreNLP toolkit served as tokenizer \cite{manning2014stanford}, while the document are truncated to a length of 500 while questions are similar truncated to 35 words as questions/passages longer than this length are few. We also adopt the following hyperparameters during training:

\begin{itemize}
     \item Learning rate = 0.001
     \item Global norm gradient clipping threshold = 5
     \item Adam optimizer with $\beta_1=0.9$, $\beta_2=0.999$
     \item Dropout = [0.1, 0.2, 0.3, 0.4, 0.5]
     \item Hidden state size =  200
\end{itemize}
 
We also experimented with trainable vs. constant embeddings and found the latter to give better validation score (1-2\% lift) due to the relatively small vocabulary in the dataset. 

\subsection{Results}


\begin{tabular}{|l r r r|}
\hline\hline
System & val EM/F1  & dev EM/F1 & test EM/F1 \\ [0.5ex] 
\hline\hline
Seq2seq Attetion Baseline & & &\\
Match LSTM Baseline &38.8/ 53.8&42.1/55.4& \\
Coattention + feedforward decoder &48.6/64.0 & 46.4/60.8 & 46.1/60.1\\
Coattention + ans-ptr & 49.7/65.2& &\\
Coattention + ans-ptr + dropout +word lookup & 56.0/69.8 & 61.6/72.3 &\textbf{61.9/72.8}\\[1ex]
\hline
\label{result_table}
\end{tabular}

\section{Discussion}

Even though the summarization engine did not effectively raise prediction accuracy in the current iteration in our model, we believe that the underlying idea is of great potential. As the current machine-based question-answering models do not perform well enough to challenge human readers, our attempt to draw inspiration from the human reading process is well-founded. Interestingly, other models performing the Q\&A task has developed similar summarization features to ours \cite{yin2016attention}, though their selection scheme is more sophisticated and thus presumably of higher quality.

As an immediate next step, we're planning to continue optimizing the summarization engine by (1) no longer limiting the summary to consists of consecutive sentences (2) develop better boundaries of selection in addition the natural sentence spans, and finally (3) incorporate semantic understanding of the text based on matching information flow to the question to better faciliate summarizing.

\bibliographystyle{unsrt}
\small{
\bibliography{ref.aux}{}
}





\end{document}